\documentclass[final]{cvpr}

\usepackage{times}
\usepackage{epsfig}
\usepackage{graphicx}
\usepackage{amsmath}
\usepackage{amssymb}
\usepackage{multirow}
\usepackage{color, colortbl}
\usepackage{amsmath,amsfonts,amsthm,bm}

\definecolor{Gray}{gray}{0.85}

\usepackage[pagebackref=true,breaklinks=true,colorlinks,bookmarks=false]{hyperref}

\def\cvprPaperID{2367} 
\def\confYear{CVPR 2021}
\makeatletter
\newcommand{\printfnsymbol}[1]{%
  \textsuperscript{\@fnsymbol{#1}}%
}

\begin{document}
\title{Reinforced Attention for Few-Shot Learning and Beyond}

\author{Jie Hong\thanks{corresponding author} $^{1,2}$, Pengfei Fang$^{1,2}$, Weihao Li$^{2}$, Tong Zhang\printfnsymbol{1}$^{3}$, Christian Simon$^{1,2}$, 
\and
Mehrtash Harandi$^{4}$, Lars Petersson$^{2}$ \\
$^{1}$Australian National University, $^{2}$Data61-CSIRO,
$^{3}$EPFL, $^{4}$Monash University \\

{\tt\small jie.hong@anu.edu.au},
{\tt\small pengfei.fang@anu.edu.au},
{\tt\small weihao.li@data61.csiro.au},
{\tt\small tong.zhang@epfl.ch},\\
{\tt\small christian.simon@anu.edu.au},
{\tt\small mehrtash.harandi@monash.edu},
{\tt\small lars.petersson@data61.csiro.au}
}
\maketitle
\thispagestyle{empty}
\pagestyle{empty}

\begin{abstract}
Few-shot learning aims to correctly recognize query samples from unseen classes given a limited number of support samples, often by relying on global embeddings of images. In this paper, we propose to equip the backbone network with an attention agent, which is trained by reinforcement learning. The policy gradient algorithm is employed to train the agent towards adaptively localizing the representative regions on feature maps over time. We further design a reward function based on the prediction of the held-out data, thus helping the attention mechanism to generalize better across the unseen classes. The extensive experiments show, with the help of the reinforced attention, that our embedding network has the capability to progressively generate a more discriminative representation in few-shot learning. Moreover, experiments on the task of image classification also show the effectiveness of the proposed design.
\end{abstract}

\section{Introduction}
The success of deep learning models rely heavily on a significant amount of labeled data, but the availability of large datasets is still limited due to the labor-intensive data preparation, which motivates the significant interest in few-shot learning \cite{koch2015siamese, ravi2016optimization, snell2017prototypical, finn2017model, sung2018learning, simon2020modulating, simon2021geodesic}. Few-shot learning aims to enable the model to classify unlabeled query examples of unseen classes, utilizing a very small number of labeled support examples. One prominent category of methods is the model-initialization based approach \cite{ravi2016optimization, finn2017model, simon2020modulating}. It temporarily updates the model parameter using support examples via gradient descent steps for the training tasks, and seeks a representation that generalizes well in the testing phase. Another line of work, the metric-learning based methods \cite{koch2015siamese, snell2017prototypical, sung2018learning}, is based on the complex manipulation of global embeddings learned by the backbone network.

\begin{figure}
\centering
	\includegraphics[width=8cm]{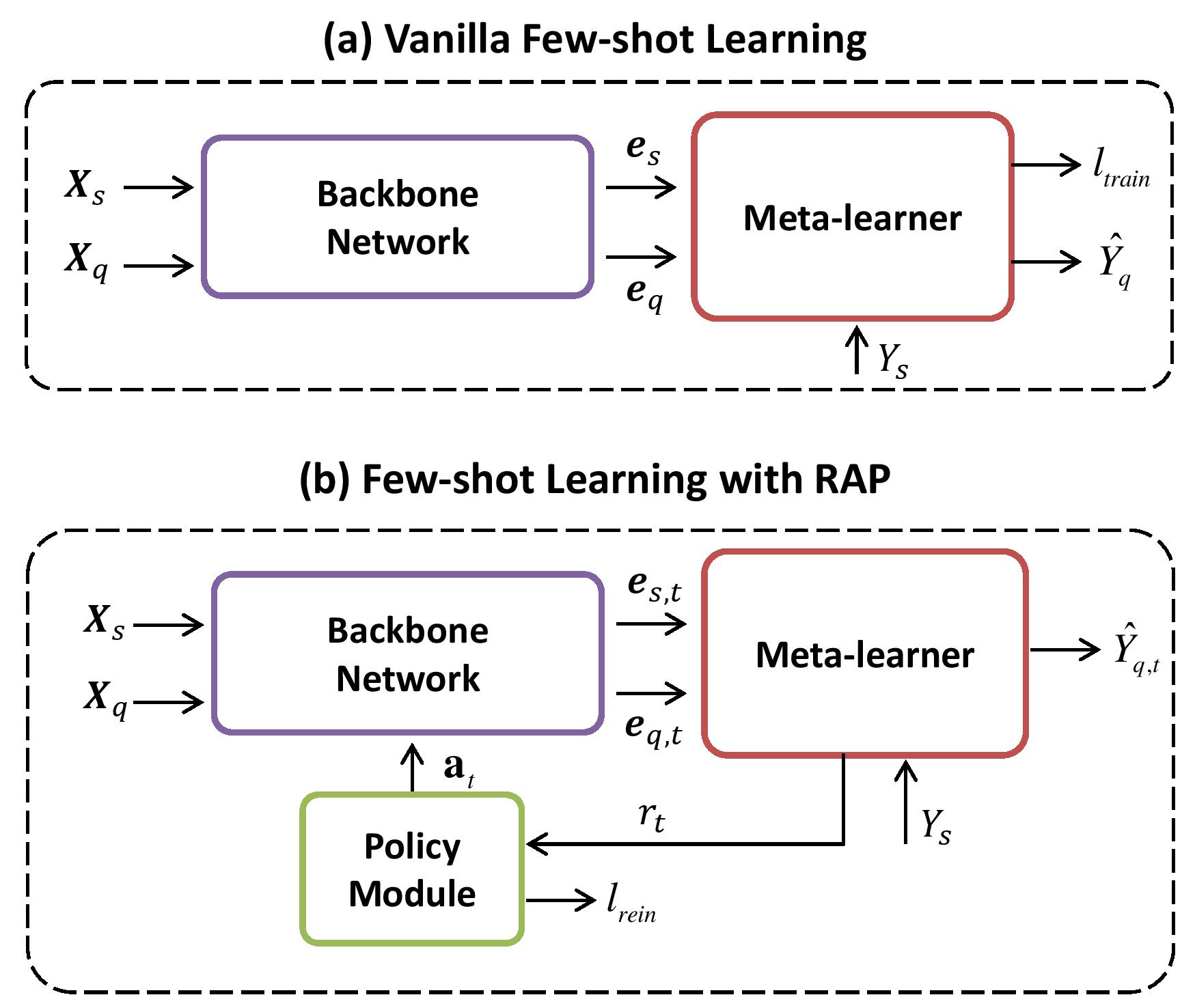}
 	\caption{Few-shot learning processes. (a) Vanilla few-shot learning. (b) Few-shot learning with RAP. The learning process has been formulated as a Markov Decision Process (MDP).}\label{rap_few}
\end{figure}

Even though the traditional approaches work well for the task of few-shot learning, they are likely to ignore the spatial information encoded within feature maps, which make the model very sensitive to the background clutter on image examples \cite{yan2019dual}. To fully make use of the available spatial information, attention-oriented designs are recently developed for few-shot learning \cite{vinyals2016matching, wang2017multi, chu2018learning, yan2019dual, ren2019incremental, hou2019cross, fan2020few}. Using word embeddings as auxiliary data, semantics-guided attention modules are proposed to capture the relevant visual features among query samples \cite{wang2017multi, chu2018learning, yan2019dual}. In addition to the semantics-guided attention, the sample-guided attention designs are able to further explore the feature relevance between support samples and query samples \cite{vinyals2016matching, ren2019incremental, hou2019cross}. While these attention models effectively make class features more representative, they tend to focus too much on designing a complex meta-learner.

In order to address the aforementioned weakness, in this work, we propose a reinforced-attention policy (RAP) model for few-shot learning, an attention mechanism trained by reinforcement learning. Specifically, an auxiliary agent is designed to equip the backbone network for computing a series of attention maps which recurrently decide where to enforce or ignore over the feature maps. 

Our designed RAP enables the backbone network to identify informative parts of the feature maps of an example, and thus make generated embeddings more discriminative for the few-shot meta-learner. We formulate the feature extraction of examples as a Markov Decision Process (MDP) and optimize RAP in a reinforcement learning setting. Given example images, the agent which progressively refines the attention upon feature maps over time is optimized according to the online feedback, \textit{i.e.}, the computed reward. The specific reward function which incorporates the performance of the meta-learner on the held-out data is designed to guide the agent towards being more generic. 

More details are specified in Fig. \ref{rap_few}. As illustrated in Fig. \ref{rap_few} (a), the vanilla few-shot learning process can be viewed as a two-component paradigm: the feature extractor, \ie the backbone network and the meta-learner. Given several query examples $\textbf{X}_{q} = \{\textbf{X}_{q,1}, \textbf{X}_{q,2}, ..., \textbf{X}_{q, i}, ... \}$ and support examples $\textbf{X}_{s} = \{ \textbf{X}_{s,1}, \textbf{X}_{s,2}, ..., \textbf{X}_{s,j}, ... \}$ with labels $Y_{s} = \{Y_{s,1}, Y_{s,2}, ..., Y_{s,j}, ... \}$, the meta-learner works as a classifier to identify the category $\hat{Y}_{q} = \{ \hat{Y}_{q,1}, \hat{Y}_{q,2}, ..., \hat{Y}_{q,i}, ... \}$ of query examples based on embedding features, $\textbf{e}_{q} = \{\textbf{e}_{q,1}, \textbf{e}_{q,2}, ..., \textbf{e}_{q,i}, ... \}$ and $\textbf{e}_{s} = \{\textbf{e}_{s,1}, \textbf{e}_{s,2}, ..., \textbf{e}_{s,j}, ... \}$, computed by the backbone network. How RAP works on a few-shot learning task is further shown in Fig. \ref{rap_few} (b). Using RAP to equip the backbone network, we convert the few-shot learning into a MDP. The policy module continuously receives the reward $r_t$ as feedback from the meta-learner and gives the action $\textbf{a}_t$ towards the larger total reward. The embedding $\textbf{e}_T$ from the last time step $T$ is viewed as the resulting embedding. Hence, instead of $\hat{Y}_q$, we take $\hat{Y}_{q,T}$  as the final prediction. The modification to only the backbone network makes RAP skip the further design of the meta-learner, such that RAP is able to be embedded in most existing few-shot learning baselines.

The contributions of this work can be summarized as follows: i$)$ Our proposed RAP is capable of attending to informative regions of feature maps while avoiding the extra cumbersome meta-learner design. Additionally, most of the few-shot learning baselines can be equipped with RAP, since RAP is essentially a flexible extension specific to the backbone network. ii$)$ We provide a novel solution to train the attention mechanism by using reinforcement learning. Intuitively, the recurrent formulation in a reinforcement learning manner can help the attention mechanism to incrementally locate useful parts of the features due to the characteristic that reinforcement learning is able to substantially learn from experience. 

In the experimental part of our few-shot learning, we select several baselines for which RAP agents are trained. In effect, our embedded design pushes the backbone network to produce embeddings which become more discriminative. Aside from few-shot learning, our design is applicable to image classification. The effectiveness is demonstrated via experiments on multiple benchmark datasets.
\section{Related Work}

\paragraph{Attention Design.} The attention mechanism, aimed at attending to the discriminative areas adaptively, have been studied extensively for 2D and 3D visual tasks~\cite{hu2018squeeze, woo2018cbam, WangXiaolong2017CVPRNonLocalNN, mnih2014recurrent, zoran2020towards, Fang_2019_ICCV, qiu2019geometric}. Those attention blocks either focus on the channel encoding~\cite{hu2018squeeze, qiu2019geometric} or spatial context connection~\cite{WangXiaolong2017CVPRNonLocalNN}. Hu \etal first propose the Squeeze-and-Excitation Network (SENet), to weight each slice of the feature map~\cite{hu2018squeeze}. Later work, termed Convolutional Block Attention Module (CBAM) \cite{woo2018cbam}, further employs hybrid spatial and channel features for attention design. In~\cite{WangXiaolong2017CVPRNonLocalNN}, the spatial context connection is established by the visual similarity between the features of the query and the key. Different from existing works, our attention design selects the informative areas by recurrently attending to the feature maps in a reinforcement learning manner. The work in ~\cite{mnih2014recurrent, zoran2020towards} also uses the recurrent model to learn an attention mask. In our work, RAP is expected to boost its generalization power by the learning experience on the held-out data.

\paragraph{Few-shot Learning.} Few-shot learning originates from the task in imitation of the human learning ability. Human beings are able to recognize a class given a few instances. Model-Agnostic Meta-Learning (MAML) \cite{finn2017model} and Prototypical Network  (ProtoNet) \cite{snell2017prototypical} are viewed as representatives for model-initialization based methods and metric-learning based methods, respectively. The former quickly adapts the classifier to the target task by learning a sensitive initialization. The latter accurately calculates the prototype of each class, and minimizes the distance in a embedding space between query samples and their corresponding prototype per class. Negative margin loss (Neg-Margin) is introduced to metric-learning based methods in \cite{liu2020negative}. Some methods, which we call graph-construction based methods, are recently studied by exploring the structural information among examples. Liu \textit{et al.} \cite{liu2018learning} develop Transductive Propagation Network (TPN) where a graph to propagate labels from labeled samples to unlabeled samples is learnt. Similar to TPN, the framework named Transfer Simplified Graph Convolutional Network (Transfer+SGC) proposed in \cite{hu2020exploiting} comes with the use of the graph convolutional network to share the information between the unlabeled examples and labeled examples. Ziko \textit{et al.} \cite{ziko2020laplacian} propose a graph clustering method (LaplacianShot) which encourages the neighboring query samples to have the same label assignment. Keeping the original meta-learner design, we extend four baseline models with RAP: MAML, ProtoNet, LaplacianShot and Neg-Margin.   

\paragraph{Attention in Few-shot Learning.} There are two main categories of attention mechanism applied in few-shot learning. The first one is semantics-guided attention \cite{wang2017multi, chu2018learning, yan2019dual, Ali_2021_CVPR}. Guided by word embeddings, Multi-attention Network \cite{wang2017multi} generates attention maps over visual features of examples, hence the representative parts of an example can be precisely captured. Like Multi-attention Network, an attention generator is developed in \cite{chu2018learning} to localize the relevant regions in an image with the help of word embeddings. Yan \textit{et al.} \cite{yan2019dual} develop a dual attention network (STANet) trained with a semantic-aware loss. The other category is sample-guided attention \cite{vinyals2016matching, ren2019incremental, hou2019cross}. Matching Network (MatchingNet) \cite{vinyals2016matching} uses an attention based on the softmax function to fully specify the prediction of the meta-learner classifier. Similar to the Matching Network, Cross Attention Network (Cross-Attention) proposed by Hou \textit{et al.} \cite{hou2019cross} aims at modeling the semantic dependency between support samples and query samples. The relevant regions on the query samples are adaptively localized such that the discrimination of embedding features benefit. MatchingNet provides attention on embeddings while Cross-Attention manipulates feature maps. Under the setting of incremental few-shot learning, an attention attractor model \cite{ren2019incremental} is presented to regulate the meta-learning of unseen classes by attending to seen classes. In contrast to these works, our developed RAP attempts to sequentially capture the important information within the feature maps of the backbone network. Its simple extension to the backbone network avoids the 
complicated meta-learner structure design. In addition, the external semantic data resource is not needed.

\paragraph{Reinforcement Learning on Visual Tasks.} Deep learning is widely investigated in the domain of reinforcement learning. By introducing the deep neural network to building the value function or the policy function, the reinforcement learning agent has a powerful capacity to learn the dynamics of the environment with which it interacts. Deep Q Network (DQN) \cite{mnih2015human} and Deep Deterministic Policy Gradient (DDPG) \cite{lillicrap2015continuous} are the very first attempts to apply the deep neural network in a reinforcement learning setting. Recently, a number of works introduce deep reinforcement learning to the computer vision community. Reinforcement learning is easily utilized in sequential visual tasks, like tracking \cite{yun2017action, chen2018real, ren2018deep}, action recognition \cite{yeung2016end, tang2018deep, chen2018part}, visual navigation \cite{wang2019reinforced, zhu2019sim}, video summarization \cite{li2018local} and person re-id \cite{lan2017deep}. The applications are also studied on more classic tasks, such as classification \cite{xu2019learning}, object detection \cite{caicedo2015active, mathe2016reinforcement}, segmentation \cite{araslanov2019actor, zhou2019context} and pose estimation \cite{krull2017poseagent, shao2020pfrl}. However, few works employing reinforcement learning have been done so far in the domain of few-shot learning. 
\section{Methodology}
\begin{figure}
\centering
	\includegraphics[width=8.5cm]{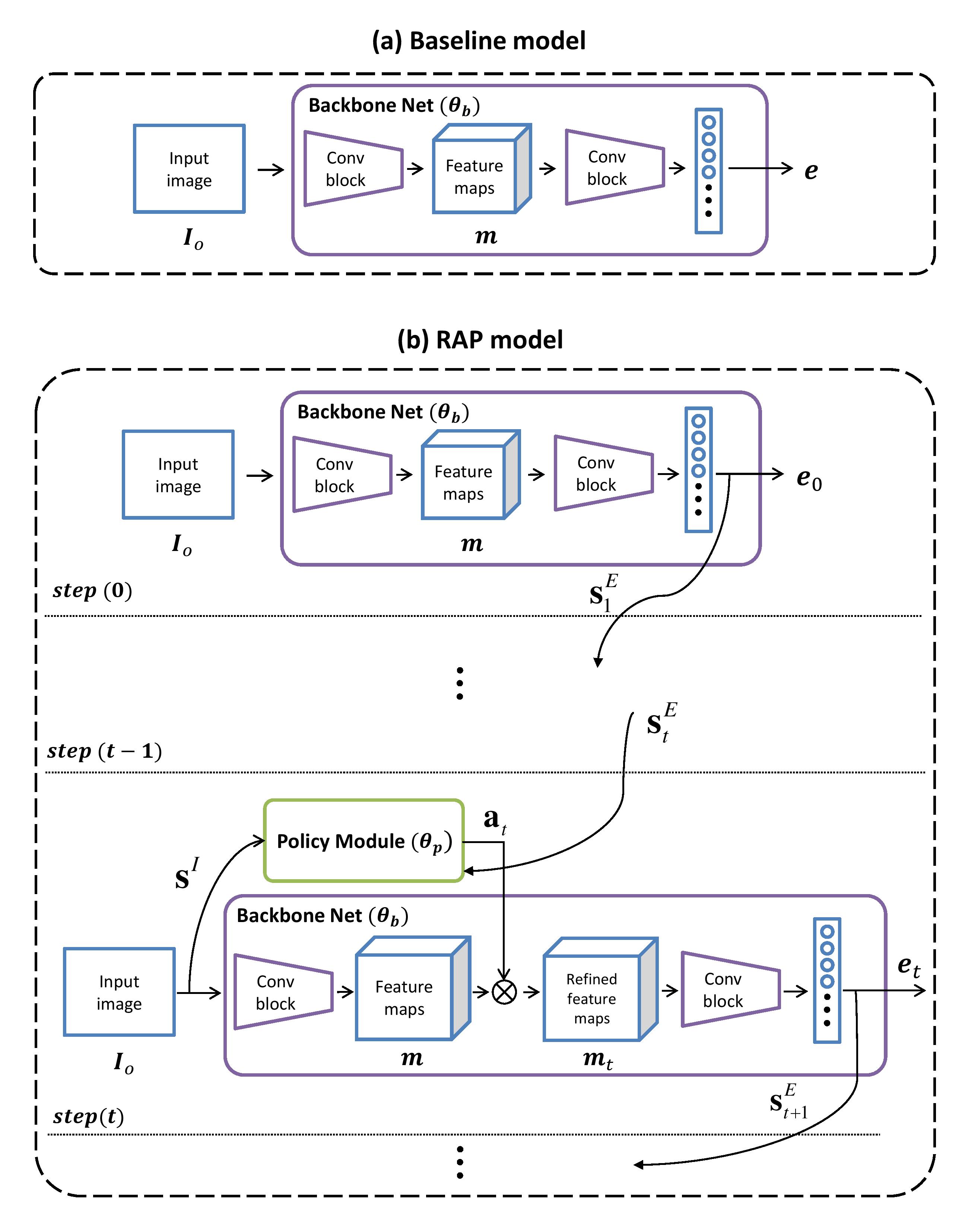}
 	\caption{The frameworks of baseline model and RAP model. (a) Baseline model. Given the input image $\textbf{I}_o$, the backbone network generates the embedding $\textbf{e}$. (b) RAP model. The recurrent formulation is built where each input image is processed over multiple time-step. The designed policy module is combined with the backbone network to compute the action $\textbf{a}_t$. The embedding computed after the final time step $\textbf{e}_{T}$ is the resulting embedding.}\label{rap}
\end{figure}

The overviews of relevant frameworks are depicted in Fig. \ref{rap}. As shown in Fig. \ref{rap} (b), the baseline model (see Fig. \ref{rap} (a)) is equipped with a policy module which receives recurrent signals and reinforces the backbone network to attend to the discriminative areas within the feature maps. Conditioned on an example image, the RAP model sequentially seeks the scoring tensor for the corresponding feature maps generated by the backbone network. During this process, we leverage reinforcement learning to train a reward-directed agent, \textit{i.e.}, the policy module (see Fig. \ref{policy}). Note that how RAP model is applied in few-shot learning is illustrated in Fig. \ref{rap_few}.  

\subsection{Problem Statement}
As described in Fig. \ref{rap} (b), one execution of RAP can be understood as a MDP. RAP agent, \textit{i.e.}, the policy module interacts with the backbone network via multiple time steps. Its behavior can be represented as state-action pairs.  At time step $t$, the agent observes the state $\textbf{s}_t$, executes the action $\textbf{a}_t$ and receives the reward $r_t$ towards optimizing the policy. The action at current time step $\textbf{a}_t$ only depends on the current state $\textbf{s}_t$ and the next state $\textbf{s}_{t+1}$ is conditioned on both $\textbf{s}_t$ and $\textbf{a}_t$.

To be specific, given feature maps from the backbone network, our RAP module will recurrently attend to informative areas within feature maps, as the RAP agent will determine attention maps until more informative areas in feature maps are found. In our setting, the action $\textbf{a}_t$ equals the attention maps. The varying factors, such as the input image $\textbf{I}_o$ and the computed embeddings $\textbf{e}_t$, are modeled as states $\textbf{s}_t$ in RL. In the following, we will introduce more details about state, action and reward.

\paragraph{State:} Under reinforcement learning settings, we normally encode the observed environment dynamics as states. As depicted in Fig. \ref{rap} (b), the dataset and backbone network are modeled as the environment to the agent. The image itself $\textbf{I}_{o} \in \mathbb{R}^{H\times W \times 3}$ and the embedding feature vector $\textbf{e}_{t} \in \mathbb{R}^{N_{SE} \times 1}$ generated from the last convolutional block are aggregated as the state $\textbf{s}_{t} = \{ \textbf{s}^{I} = \textbf{I}_{o}, \textbf{s}^{E}_{t} = \textbf{e}_{t} \}$ at time step $t \in \{0, 1, 2, ..., T\}$. $T$ is the total execution times to each image. It is noted that $\textbf{s}^{I}$ is fixed in one sequence of processing. We need to address the issue that the high variance exists among tremendous images \cite{mnih2014recurrent, mao2018variance}. For this purpose, we treat the image $\textbf{I}_{o}$ as one of states $\textbf{s}^{I}$. Hence,  RAP is capable of learning the variance among input images. In addition, the setting of the state enables the agent to fuse both the low-level information and the high-level information. The high-level information from $\textbf{s}^{E}_{t}$ indicates the how action affects the backbone network. The low-level information which is encoded within $\textbf{s}^{I}$ represents the image variance.

\paragraph{Action:} Attention computed at time $t$ for weighting each pixel on feature map are defined as the action $\textbf{a}_{t} \in \mathbb{R}^{h \times w \times c}$ where $h$, $w$ and $c$ are its height, width and length. As illustrated in Fig. \ref{policy}. the action before reshaping operation $\textbf{a}^{v}_{t} \in \mathbb{R}^{hw \times 1}$ is drawn from the distribution $\pi$: $\textbf{a}^{v}_{t} \sim \pi(.|g(\textbf{s}_{t} | \bm{\theta}_{p}))$ where $\pi$ is the distribution function parameterized by the computation from policy function $g$. Then, $\textbf{a}^{v}_{t} \in \mathbb{R}^{hw \times 1}$ is reshaped and duplicated to $\textbf{a}_{t} \in \mathbb{R}^{h \times w \times c}$ which has the same size as the operated feature map. The element-wise multiplication is performed to give the refined feature map as follows:
\begin{equation}
\textbf{m}_{t} = \textbf{a}_{t} \otimes \textbf{m}.
\end{equation}
where $\textbf{m} \in \mathbb{R}^{h \times w \times c}$ ($\textbf{m}_{t} \in \mathbb{R}^{h \times w \times c}$) is the feature map before (after) the action being taken at time $t$. Given the state $\textbf{s}_{t}$, the policy function $g$ can decide which pixel along the channel in the feature map should be heightened or weakened at time $t$.

\paragraph{Reward:} After the execution of the action $\textbf{a}_{t}$ over feature maps at step $t$, the agent receives the reward $r_{t}$ which evaluates $\textbf{a}_{t}$. The reward function matters in the sense that it criticizes the direction that the policy is optimized towards. One goal of introducing RAP is to adaptively allocate the informative areas over feature maps via a sequential process. In doing so, one option is to closely monitor and react to the model performance on the held-out data. This can be achieved with the help of the validation set. Hence, the reward function is proposed as 
\begin{equation}
\mathbf{r}_{t} = - \alpha \ell_{val, t} \label{reward}
\end{equation} 
where $\ell_{val, t}$ is the loss built on the validation set at time $t$ and $\alpha$ is the coefficient. In our case, the reward design aims at directing the policy to achieve the higher prediction on the validation data. By explicitly pursuing the better performance on the validation data, RAP is able to have the better generalization ability and therefore pay attention to more useful information on feature maps. 

\subsection{Policy Design}
\begin{figure}
\centering
	\includegraphics[width=8.5cm]{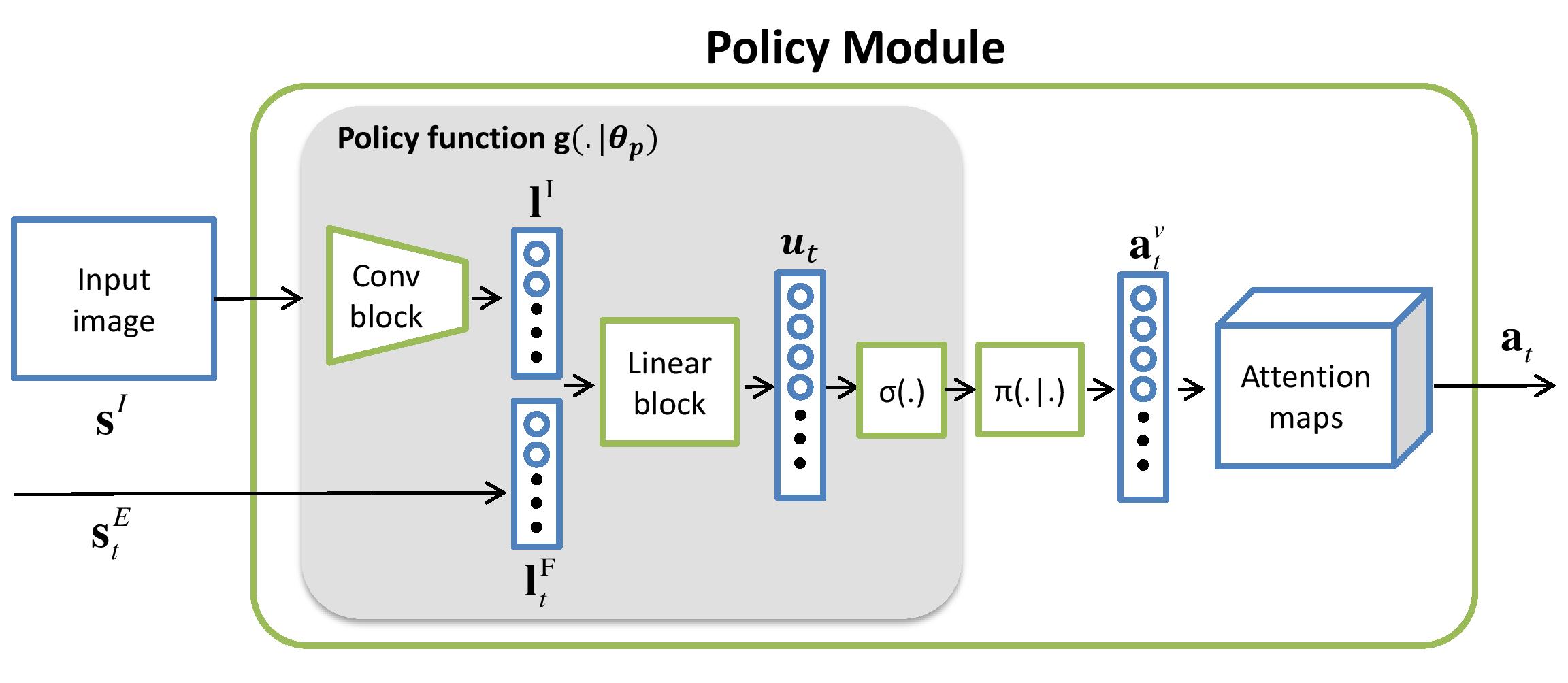}
 	\caption{The framework of the policy module. The module essentially maps the state $\textbf{s}_t$ to the action $\textbf{a}_t$. $\sigma(.)$ is Sigmoid function.}\label{policy}
\end{figure}

In this case, we exploit reinforcement learning to learn the policy. The architecture of the policy module is illustrated in Fig. \ref{policy}. Let $g(\textbf{s}_{t}|\bm{\theta}_{p})$ be the policy function which outputs the vector $\textbf{u}_{t} \in \mathbb{R}^{hw \times 1}$, representing the parameter value of the distribution function $\pi(.|g(. | \bm{\theta}_{p}))$. Under this distribution, $\textbf{a}^{v}_{t} \in \mathbb{R}^{hw \times 1}$ can be chosen.

As shown in Fig. \ref{policy}, given the image $\textbf{s}^I$, the conv block, which contains standard conv operations followed by the global average pooling along each channel, is used to compute a $N_{l1}$-dimensional representation $\textbf{l}^{I} \in \mathbb{R}^{N_{lI} \times 1}$. The resulting vector $\textbf{l}^{I}$ is then concatenated with another state vector $\textbf{l}^{E}_{t} = \textbf{s}^{E}_{t} \in \mathbb{R}^{N_{lE} \times 1}$ passed from the last time step to form the vector $\textbf{l}_{t} = \{ \textbf{l}^{I}, \textbf{l}^{E}_{t} \}$ which encodes the low-level spatial information from the given example image and the high-level semantic information from the backbone network. Taking $\textbf{l}_{t}$ as input, the linear block follows to calculate $\textbf{u}_{t}$. The configuration of the linear block is constructed with fully connected (FC) layers. The computed action vector $\textbf{a}^{v}_{t}$ should be reshaped and extended $c$ times to $\textbf{a}_{t} \in \mathbb{R}^{h\times w \times c}$ where $c$ indicates the channel number of the feature map to be operated on. We need to avoid the situation that the proposed auxiliary mechanism itself contributes too much to boosting the performance, for which the designed convolution block is usually much shallower than the first conv block of the backbone network before the operation on the feature map.

\subsection{Policy Training}
The learnable parameter $\bm{\theta} = \{\bm{\theta}_{b}, \bm{\theta}_{p}\}$ contains the parameter of the backbone network $\bm{\theta}_{b}$ and the policy module $\bm{\theta}_{p}$. The objective of policy training is to maximize the sum of the reward $J^{\pi} = \sum^{T}_{t} r_{t}$ under the distribution $\pi$ given the input image $\textbf{s}^I$. We directly write the reinforcement loss $\ell_{rein}$ as follows:
\begin{equation}
\begin{aligned}
\ell_{rein} &= -\frac{1}{NT} \sum^{N}_{i=1} J^{\pi}_{i} \\
            &= -\frac{1}{NT} \sum^{N}_{i=1}\sum^{T}_{t=1} \mathrm{log} (\pi(\textbf{u}_{i, t}|\textbf{s}_{i, t}, \bm{\theta}_{p}) ) r_{i, t}
\end{aligned}
\end{equation}
where $\textbf{u}_{t} = g(\textbf{s}_{t} | \bm{\theta}_{p})$ on the current state $\textbf{s}_{t}$ is the predicted parameter (\textit{e.g.}, mean) to the distribution function $\pi$ in choosing $\textbf{a}^{v}_{t}$ and $\mathrm{log}\pi$ is the log-probability. $N$ is the batch size of images that the model processes in each time step.  Using the REINFORCE algorithm \cite{williams1992simple}, the gradient can be expressed as:
\begin{equation}\label{policy_gradient}
\nabla_{\theta}J^{\pi} = \sum^{T}_{t=1} \nabla_{\bm{\theta}} \mathrm{log}\big( \pi(\textbf{u}_{t}|\textbf{s}_{t}; \bm{\theta}_{p})\big) r_{t},
\end{equation}

The gradient $\nabla_{\theta}J^{\pi}$ in (\ref{policy_gradient}) encourages the optimal parameter to produce a larger reward. The parameter $\theta_{p}$ determines the sensitivity of the policy to the reward $r_{t}$. Its motivation is to seek a suitable scale acting on the feedback from the validation. Suppose $r_{i, t}$ becomes zero, the gradient $\nabla_{\theta}J^{\pi}$ likewise becomes zero, which would not push the weight to be optimized. One possible concern is that only a reinforcement learning loss $\ell_{rein}$ that does not include the train data would make the model more firmly rooted in the validation data and thus negatively degrades the accuracy. Incorporating the loss based on the train set can explicitly avoid this problem. It is therefore worth considering a composite form of the total loss as:
\begin{equation}
\ell_{total} = \ell_{rein} + \ell_{train}
\end{equation}
where $\ell_{train}$ is the loss built on the train data. When the training is under progress with $\ell_{total}$, $\bm{\theta}$ is free to be updated. The train loss $\ell_{train}$ based on the train data guarantees the basic performance of the backbone network whereas the reinforce loss $\ell_{rein}$ based on the validation data enables the policy module to actively identify the informative areas on the feature map of the backbone. When it comes to the inference stage, $\bm{\theta}$ becomes frozen.
\section{Experiment}
The proposed design is mainly evaluated on the task of few-shot learning. We assess our RAP model over the \emph{mini}ImageNet \cite{ravi2016optimization} and Caltech-UCSD Birds (CUB-200-2011) \cite{WahCUB_200_2011} datasets where it consistently outperforms the corresponding baseline model. Then, to verify that the feasibility of our design is not restricted to few-shot learning alone, experiments on image classification are performed on CIFAR10/100 and STL-10. The standard data augmentations, \textit{i.e.}, random crop and random horizontal flips, are applied to the training samples throughout all experiments. The Adam optimizer \cite{kingma2015adam} is employed to optimize the model weight. Moreover, the trained model that achieves the best validation accuracy is selected to do the inference. All deep models are implemented using PyTorch \cite{paszke2019pytorch}. We opt for Gaussian function as distribution function $\pi$ whose mean value is computed by policy function $g$.

\subsection{Datasets}
\paragraph{\emph{mini}ImageNet.} For few-shot learning, \emph{mini}ImageNet \cite{ravi2016optimization} is the most popular benchmark (60000 images of size 84$\times$84 selected from ImageNet) which is comprised of 64 training classes, 16 validation classes and 20 test classes of images. Every class has 600 images. The size ratio of the train set and validation set is 4:1.

\paragraph{CUB-200-2011.} The CUB-200-2011 dataset \cite{WahCUB_200_2011} consists of 11788 images of 84$\times$84 from 200 classes. Following the data preparation from \cite{hilliard2018few}, we randomly divide it into 100 train classes, 50 validation classes and 50 test classes.

\paragraph{CIFAR10/100.} CIFAR10/100 \cite{krizhevsky2009learning} comprises 60000 images from 10 classes/100 classes. The size of each image is $32\times32$. There are originally 50000 train images and 10000 test images. Our method needs a validation set, thus, we randomly divide the original train set into 40000 train samples and 10000 validation samples.

\paragraph{STL-10.} STL-10 \cite{coates2011analysis} contains 10 classes where each class has 500 train images and 800 test images of $96\times96$. Similar to CIFAR, the original train data is proportionately split into train data and validation data by a ratio of 4:1. 

\subsection{Few-shot Learning}
Training a few-shot learning machine is characterized as an episode training. An episode, which is randomly sampled from training data, is a simple $N$-way $K$-shot $Q$-query classification task. The training process is to classify $Q \times N$ query samples from $N$ classes correctly given $K$ samples per class (\textit{e.g.}, $K = 1$ or $5$).

The 1-shot and 5-shot scenarios are examined. Our entire set of experiments are performed using the setting of 5-way and 16-Query. Moreover, we arrange 1200 episodes at random for MAML and ProtoNet (10000 and 600 episodes for LaplacianShot and Neg-Margin, respectively). The final performance is provided averaging the classification accuracy over episodes. For the design details of the policy module, the conv block includes three convolutional layers with 3$\times$3 convolutions where each layer is followed by a batch normalization, ReLU and max pooling. One-layer FC is following as a linear block. The total time step $T$ is kept fixed as 5. The coefficient $\alpha$ in (\ref{reward}) is set to 1e-4. Different architectures of backbone network are assessed including Conv-4 \cite{vinyals2016matching}, Conv6, ResNet-10, ResNet-12, ResNet-18 \cite{he2016deep} and DenseNet-121 \cite{huang2017densely}. We insert our attention module after the second (or third) conv block of Conv-4 and Conv-6 (or ResNet-10, ResNet-12, ResNet-18 and DenseNet-121).


\subsubsection{\emph{mini}ImageNet}
In the first part of our few-shot learning experiments, we compare the performance on \emph{mini}ImageNet. The results are presented in Table \ref{table_mini}. We equip the classic baselines including MAML and ProtoNet with our proposed RAP. LaplacianShot, an excellent baseline, is also considered. In most cases, our RAP models outperform their corresponding baseline counterparts except for RAP-LaplacianShot with DenseNet-121 in the 5-way 1-shot setting. Another observation is that the choice of baseline do affect the performance improvement against baselines. RAP-MAML and RAP-ProtoNet yield a greater performance improvement than RAP-LaplacianShot. Comparisons with other few-shot learning baselines which exploit attention are provided in Table \ref{table_mini_attention}. As we can see from the table, our performance is better than others given the same backbone. 

\begin{table}
\begin{center}
\resizebox{.47\textwidth}{!}{
\begin{tabular}{c|ccccc}
\hline

&   &\multicolumn{2}{c}{\textbf{5-way Acc.}}   \\  \hline

\textbf{Model} &\textbf{Backbone} &\textbf{1-shot} &\textbf{5-shot}  \\ \hline\hline

\multirow{3}*{MAML \cite{finn2017model}} 
&Conv-4     &48.70$\pm$1.84  &63.15$\pm$0.92  \\

&Conv-6     &50.96$\pm$0.92  &66.09$\pm$0.71  \\

&ResNet-10  &54.69$\pm$0.89  &66.62$\pm$0.83  \\ 

\rowcolor{Gray}  
&Conv-6     &52.57$\pm$0.61   &66.96$\pm$0.52     \\ 

\rowcolor{Gray}\multirow{-2}*{\textbf{RAP-MAML}} 
&ResNet-10  &\textbf{56.13$\pm$0.62}  &\textbf{68.74$\pm$0.54}     \\ \hline

\multirow{3}*{ProtoNet \cite{snell2017prototypical}} 
&Conv-4     &49.42$\pm$0.78  &68.20$\pm$0.66  \\

&Conv-6     &50.37$\pm$0.83  &67.33$\pm$0.67  \\

&ResNet-10  &51.98$\pm$0.84  &72.64$\pm$0.64  \\

\rowcolor{Gray} 
&Conv-6     &51.72$\pm$0.72  &69.18$\pm$0.45      \\

\rowcolor{Gray}\multirow{-2}*{\textbf{RAP-ProtoNet}}  
&ResNet-10  &\textbf{53.64$\pm$0.60}  &\textbf{74.54$\pm$0.45}  \\ \hline

TPN \cite{liu2018learning}
&ResNet-12  &59.46  &75.65  \\

LTS \cite{li2019learning}
&ResNet-12  &70.1$\pm$1.9  &78.7$\pm$0.8  \\

MetaOptNet-SVM \cite{lee2019meta}
&ResNet-12  &64.09$\pm$0.62  &80.00$\pm$0.45  \\

Neg-Margin \cite{liu2020negative}
&ResNet-12  &63.85$\pm$0.81  &81.57$\pm$0.56  \\

DSN-MR \cite{simon2020adaptive}
&ResNet-12  &67.09$\pm$0.68  &81.65$\pm$0.69  \\

Su \etal~\cite{su2019does} &ResNet-18 &- &76.60$\pm${0.70} \\

Hyperbolic ProtoNet~\cite{khrulkov2020hyperbolic} &ResNet-18 &59.47$\pm${0.20} &76.84$\pm${0.14} \\

LwoF~\cite{gidaris2018dynamic} &WRN &60.06$\pm$0.14  &76.39$\pm$0.11 \\  

wDAE-GNN~\cite{gidaris2019generating} &WRN&61.07$\pm${0.15} &76.75$\pm$0.11 \\

EPNet \cite{rodriguez2020embedding}
&WRN  &70.74$\pm$0.85  &84.34$\pm$0.53  \\ 

\hline
\multirow{3}*{LaplacianShot \cite{ziko2020laplacian}} 
&ResNet-10*   &69.47$\pm$0.20  &79.78$\pm$0.15  \\

&ResNet-12*    &72.29$\pm$0.20  &82.85$\pm$0.14  \\

&DenseNet-121  &75.57$\pm$0.19  &84.72$\pm$0.13  \\

\rowcolor{Gray}
&ResNet-10     &71.34$\pm$0.19  &81.98$\pm$0.14  \\

\rowcolor{Gray}
&ResNet-12     &74.29$\pm$0.20  &84.51$\pm$0.13  \\

\rowcolor{Gray}\multirow{-3}*{\textbf{RAP-LaplacianShot}} 
&DenseNet-121  &\textbf{75.58$\pm$0.20}  &\textbf{85.63$\pm$0.13}  \\ \hline

\end{tabular}} 
\end{center}
\caption{Few-shot classification accuracy on \emph{mini}ImageNet. Published results of MAML and ProtoNet with Conv4 are provided in \cite{finn2017model} and \cite{snell2017prototypical}, respectively while their outcomes under Conv-6 and ResNet-10 are reported in \cite{chen2019closerfewshot}. "*" indicates the result is obtained using networks implemented by us.} \label{table_mini}
\end{table}

\begin{table}
\begin{center}
\resizebox{.47\textwidth}{!}{
\begin{tabular}{c|ccccc}
\hline

&   &\multicolumn{2}{c}{\textbf{5-way Acc.}}   \\  \hline

\textbf{Model} &\textbf{Backbone} &\textbf{1-shot} &\textbf{5-shot}  \\ \hline\hline

Attention Attractor \cite{ren2019incremental}
&ResNet-10    &54.95$\pm$0.30  &63.04$\pm$0.30  \\

\rowcolor{Gray} \textbf{RAP-MAML}
&ResNet-10     &56.13$\pm$0.62  &68.74$\pm$0.54  \\

\rowcolor{Gray} \textbf{RAP-ProtoNet}
&ResNet-10     &53.64$\pm$0.60  &74.54$\pm$0.45  \\

\rowcolor{Gray} \textbf{RAP-LaplacianShot}
&ResNet-10     &\textbf{71.34$\pm$0.19}  &\textbf{81.98$\pm$0.14}  \\ \hline

STANet \cite{yan2019dual}
&ResNet-12    &58.35$\pm$0.57  &71.07$\pm$0.39  \\

Cross-Attention \cite{hou2019cross}
&ResNet-12    &67.19$\pm$0.55  &80.64$\pm$0.35  \\

\rowcolor{Gray} \textbf{RAP-LaplacianShot}
&ResNet-12     &\textbf{74.29$\pm$0.20}  &\textbf{84.51$\pm$0.13}  \\ \hline

\end{tabular}} 
\end{center}
\caption{Comparison with other attention designs applied to few-shot learning on \emph{mini}ImageNet. Different attention models are compared under the same backbone.} \label{table_mini_attention}
\end{table}

\subsubsection{CUB-200-2011}
Below, experiments are performed on CUB-200-2011. In this case, besides MAML and ProtoNet and LaplacianShot, we test another strong baseline, Neg-Margin. The obtained accuracy are reported in Table \ref{table_cub}. It is obvious that using MAML, RAP models have an accuracy increase of around 2\% to 6\% against the baseline models whatever the choice of backbone. Similarly, it is also the case for ProtoNet, LaplacianShot and Neg-Margin, that the performance of RAP models are significantly superior to the corresponding baselines. The competitive outcomes prove that the learned policy by reinforcement learning helps the backbone network to localize the informative areas of the feature map and thus generate discriminative embeddings. It should also be emphasized, again, that our RAP is compatible with a number of various few-shot learning baselines.

\begin{table}
\begin{center}
\resizebox{.47\textwidth}{!}{
\begin{tabular}{c|ccccc}
\hline

&   &\multicolumn{2}{c}{\textbf{5-way Acc.}}   \\  \hline

\textbf{Model} &\textbf{Backbone} &\textbf{1-shot} &\textbf{5-shot}  \\ \hline\hline

\multirow{4}*{MAML \cite{finn2017model}} 
&Conv-4     &54.73$\pm$0.97  &75.75$\pm$0.76  \\

&Conv-6     &66.26$\pm$1.05  &78.82$\pm$0.70  \\

&ResNet-10  &70.32$\pm$0.99  &80.93$\pm$0.71  \\ 

&ResNet-18  &68.42$\pm$1.07  &83.47$\pm$0.62  \\ 

\rowcolor{Gray}  
&Conv-4    &61.49$\pm$0.70   &77.15$\pm$0.50     \\ 

\rowcolor{Gray}  
&Conv-6    &69.95$\pm$0.68   &81.48$\pm$0.44     \\ 

\rowcolor{Gray}
&ResNet-10  &74.33$\pm$0.65   &83.29$\pm$0.42 \\

\rowcolor{Gray}\multirow{-4}*{\textbf{RAP-MAML}} 
&ResNet-18  &\textbf{75.04$\pm$0.70}   &\textbf{86.07$\pm$0.43}     \\ \hline

\multirow{4}*{ProtoNet \cite{snell2017prototypical}} 
&Conv-4     &50.46$\pm$0.88  &76.39$\pm$0.64  \\

&Conv-6     &66.36$\pm$1.00  &82.03$\pm$0.59  \\

&ResNet-10  &73.22$\pm$0.92  &85.01$\pm$0.52  \\

&ResNet-18  &72.99$\pm$0.88  &86.64$\pm$0.51  \\

\rowcolor{Gray} 
&Conv-4     &56.71$\pm$0.66  &78.70$\pm$0.44      \\

\rowcolor{Gray} 
&Conv-6     &67.79$\pm$0.66  &83.78$\pm$0.41      \\

\rowcolor{Gray} 
&ResNet-10  &\textbf{75.17$\pm$0.63}      &88.29$\pm$0.34  \\ 

\rowcolor{Gray}\multirow{-4}*{\textbf{RAP-ProtoNet}}  
&ResNet-18  &74.09$\pm$0.60      &\textbf{89.23$\pm$0.31}  \\ \hline

DeepEMD~\cite{zhang2020deepemd}         &ResNet-12 &76.65$\pm${0.83}         &88.69$\pm${0.50} \\

MatchingNet~\cite{vinyals2016matching}  &ResNet-18  &72.36$\pm$0.90 &83.64$\pm$0.60  \\

RelationNet~\cite{sung2018learning}     &ResNet-18  &67.59$\pm$1.02          &82.75$\pm$0.58 \\

Chen \etal~\cite{chen2019closerfewshot} &ResNet-18 &67.02 & 83.58 \\

SimpleShot~\cite{wang2019simpleshot}    &ResNet-18 &70.28 &86.37 \\

Manifold \cite{mangla2020charting}
&WRN  &80.68$\pm$0.81  &90.85$\pm$0.44  \\ 

EPNet \cite{rodriguez2020embedding}
&WRN  &87.75$\pm$0.70  &94.03$\pm$0.33  \\ \hline

Neg-Margin \cite{liu2020negative}
&ResNet-18     &72.66$\pm$0.85  &89.40$\pm$0.43  \\

\rowcolor{Gray}\textbf{RAP-Neg-Margin}
&ResNet-18     &\textbf{75.37$\pm$0.81}  &\textbf{90.61$\pm$0.39}  \\ \hline

LaplacianShot \cite{ziko2020laplacian}
&ResNet-18     &80.96   &88.68  \\

\rowcolor{Gray}\textbf{RAP-LaplacianShot}
&ResNet-18     &\textbf{83.59$\pm$0.18}  &\textbf{90.77$\pm$0.10}  \\ \hline

\end{tabular}} 
\end{center}
\caption{Few-shot classification accuracy on CUB-200-2011. The results of MAML and ProtoNet are provided from \cite{chen2019closerfewshot}.} \label{table_cub}
\end{table}

\subsection{Image Classification}
For the task of image classification, all models are trained with a mini-batch size of 128 for 4000 epochs (2000 epochs on STL10). Pre-trained backbones are used. The learning rate is set to 1e-6. The RAP models execute actions for $T$=5 times. The results of the classification accuracy on three datasets are listed in Table \ref{table_cifar} where RAP models surpass the corresponding baseline models with a healthy margin. We can observe that on each dataset, RAP models have a similar accuracy improvement given different backbone architectures.

\begin{table}
\begin{center}
\resizebox{.40\textwidth}{!}{
\begin{tabular}{c|c|cc} 
\hline

\textbf{Dataset}  &\textbf{Model}  &\textbf{Backbone}  &\textbf{Accuracy (\%)}    \\ \hline\hline

\multirow{8}*{\textbf{CIFAR10}}
&\multirow{4}*{Baseline}  
        &LeNet*     & 75.94         \\
&       &VGG-16*    & 91.48         \\ 
&       &ResNet-18* & 92.68         \\
&       &ResNet-50* & 93.02         \\

&      &LeNet       & 77.15   ($\uparrow$ 1.21)   \\
&      &VGG-16      & 92.05 ($\uparrow$ 0.57)     \\
&      &ResNet-18   & 93.25 ($\uparrow$ 0.57)     \\
& \multirow{-4}*{\textbf{RAP}}
                      &ResNet-50   & \textbf{93.56 ($\uparrow$ 0.54)}    \\ \hline
                      
\multirow{8}*{\textbf{CIFAR100}}
&\multirow{4}*{Baseline}  
        &LeNet*        & 41.63         \\
&       &VGG-16*       & 67.23         \\ 
&       &ResNet-18*    & 72.56         \\
&       &ResNet-50*    & 74.22         \\

&      &LeNet       & 43.02 ($\uparrow$ 1.39)     \\
&      &VGG16       & 68.45 ($\uparrow$ 1.22)     \\
&      &ResNet-18   & 73.76 ($\uparrow$ 1.20)     \\
& \multirow{-4}*{\textbf{RAP}}
                      &ResNet-50   & \textbf{75.36 ($\uparrow$ 1.14)}    \\ \hline

\multirow{6}*{\textbf{STL10}}
&\multirow{3}*{Baseline}  
        &VGG-16*    & 74.43         \\
&       &ResNet-18* & 78.21         \\
&       &ResNet-34* & 76.31         \\

&      &VGG-16       & 76.16 ($\uparrow$ 1.73)     \\
&      &ResNet-18    &\textbf{80.15 ($\uparrow$ 1.94)}     \\
& \multirow{-3}*{\textbf{RAP}}  &ResNet-34   & 77.74 ($\uparrow$ 1.43)  \\  \hline      
\end{tabular}}
\end{center}
\caption{Image classification accuracy of all models on CIFAR10/CIFAR100 and STL10. LeNet \cite{lecun1998gradient}, VGG-16 \cite{simonyan2014very}, ResNet-18, ResNet34 and ResNet-50 are chosen as backbones. Baseline models are trained on both train set and validation set. "*" indicates the result is obtained using networks implemented by us.} \label{table_cifar}
\end{table}

\subsection{Analysis}
\paragraph{Validation Set.} RAP models trained by $\ell_{total} = \ell_{train} + \ell_{rein}$ work under the specific setting that $\ell_{train}$ and $\ell_{rein}$ are built on train data and validation data, respectively. Any batch is formed by images from either the training set or the validation set. The reward $r_{t}$ of RAP incorporates the feedback from the validation loss $\ell_{val}$, hence, to justify the comparison between RAP models and baseline models, it is necessary to train the baseline models by using both the train set and the validation set. Thus, the loss $\ell_{total} = \ell_{train}$ to the baseline model is replaced by $\ell_{total} = \ell_{train} + \ell_{val}$. As reported in Table \ref{table_vali}, even under the same data settings, RAP models still beat baseline models with different backbones. Another fact is that RAP models are sensitive to $\alpha$. $\ell_{rein}$ does contribute even if $\alpha$ is set to 1e-4.  We thus have to choose $\alpha$ carefully since $\alpha$ decides what extend validation data should be leveraged (see Table \ref{table_alpha}).

\begin{table}
\begin{center}
\resizebox{.45\textwidth}{!}{
\begin{tabular}{c|lcccc}
\hline

&   &\multicolumn{2}{c}{\textbf{5-way Acc.}}   \\  \hline

\textbf{Model} &\textbf{Backbone} &\textbf{1-shot} &\textbf{5-shot}  \\ \hline\hline

\multirow{2}*{MAML \cite{finn2017model}} 
&Conv-4*     &57.55$\pm$0.67   &75.61$\pm$0.49   \\

&Conv-6*     &65.07$\pm$0.72   &79.89$\pm$0.48   \\ 

\rowcolor{Gray}  
&Conv-4    &61.49$\pm$0.70   &77.15$\pm$0.50     \\  

\rowcolor{Gray}\multirow{-2}*{\textbf{RAP-MAML}} 
&Conv-6    &69.95$\pm$0.68   &81.48$\pm$0.44      \\ \hline

\multirow{2}*{ProtoNet \cite{snell2017prototypical}} 
&Conv-4*     &51.61$\pm$0.65  &75.29$\pm$0.48   \\

&Conv-6*     &64.72$\pm$0.72  &81.35$\pm$0.43   \\

\rowcolor{Gray} 
&Conv-4     &56.71$\pm$0.66  &78.70$\pm$0.44      \\

\rowcolor{Gray}\multirow{-2}*{\textbf{RAP-ProtoNet}}  
&Conv-6     &67.79$\pm$0.66  &83.78$\pm$0.41  \\ \hline

\end{tabular}} 
\end{center}
\caption{Fair comparisons on CUB-200-2011 where baseline models are trained on both the train set and validation set like RAP models. "*" indicates the result is obtained using networks implemented by us.} \label{table_vali}
\end{table}

\begin{table}
\begin{center}
\resizebox{.47\textwidth}{!}{
\begin{tabular}{c|c|cccccc}
\hline

&\multicolumn{1}{c}{}     &  &\multicolumn{2}{c}{\textbf{5-way Acc.}}   \\  \hline

\textbf{Model} &$\alpha$  &\textbf{Backbone} &\textbf{1-shot} &\textbf{5-shot}  \\ \hline\hline

\rowcolor{Gray}&0         &Conv-4 &53.83$\pm${0.66}   &77.10$\pm${0.46}      \\

\rowcolor{Gray}&1e-4      &Conv-4 &56.71$\pm${0.66}   &78.70$\pm${0.44}      \\

\rowcolor{Gray}&1e-2      &Conv-4 &44.68$\pm${0.61}   &71.02$\pm${0.49}      \\ 

\rowcolor{Gray}\multirow{-4}*{\textbf{RAP-ProtoNet}} 
&1                        &Conv-4 &44.90$\pm${0.63}   &65.07$\pm${0.50}      \\ \hline

\rowcolor{Gray}&0         &ResNet-10 &72.71$\pm${0.61}   &86.68$\pm${0.33}       \\

\rowcolor{Gray}&1e-4      &ResNet-10 &75.17$\pm${0.63}   &88.29$\pm${0.34}       \\

\rowcolor{Gray}&1e-2      &ResNet-10 &72.61$\pm${0.64}   &85.95$\pm${0.37}        \\ 

\rowcolor{Gray}\multirow{-4}*{\textbf{RAP-ProtoNet}} 
&1                        &ResNet-10 &66.33$\pm${0.64}   &80.47$\pm${0.42}         \\ \hline

\end{tabular}}
\end{center}
\caption{The results of 5-way classification of RAP-ProtoNet on CUB-200-2011 using different $\alpha$.} \label{table_alpha}
\end{table}

\vspace*{-6pt}
\paragraph{Attention Design.} In order to further verify the advantage of our method, we compare RAP against two conventional attention designs, SENet and CBAM. The policy module of RAP is substituted by different attention mechanisms, but the remaining settings are kept the same. As can be seen in Table \ref{table_attention}, models equipped with SENet and CBAM are shown to outperform the corresponding baseline models. However, their performance do not exceed our RAP models except that they have a nearly identical accuracy in the 5-way 1-shot classification using the Conv-4 backbone.

\begin{table}
\begin{center}
\resizebox{.44\textwidth}{!}{
\begin{tabular}{c|ccc}
\hline

& &\multicolumn{2}{c}{\textbf{5-way Acc.}}  \\  \hline

&\textbf{Backbone} &\textbf{1-shot} &\textbf{5-shot}\\ \hline\hline

\multirow{2}*{SENet \cite{hu2018squeeze}}
&Conv-4*     &55.18$\pm$0.61  &77.07$\pm$0.46  \\
&ResNet-10*  &73.89$\pm$0.57  &87.32$\pm$0.38  \\ \hline

\multirow{2}*{CBAM \cite{woo2018cbam}}
&Conv-4*     &55.52$\pm$0.55  &76.80$\pm$0.34  \\
&ResNet-10*  &72.43$\pm$0.65  &86.70$\pm$0.48  \\ \hline

\rowcolor{Gray} &Conv-4     &56.71$\pm$0.66  &78.70$\pm$0.44 \\
\rowcolor{Gray}\multirow{-2}*{\textbf{RAP}}
&ResNet-10                  &75.17$\pm$0.63  &88.29$\pm$0.34 \\ \hline

\end{tabular}}
\end{center}
\caption{Few-shot classification accuracy using ProtoNet as a baseline with different regular attention designs on CUB-200-2011. "*"
indicates the result is obtained using networks implemented by us.} \label{table_attention}
\end{table}

\begin{figure}
\begin{center}
	\includegraphics[width=7.5cm]{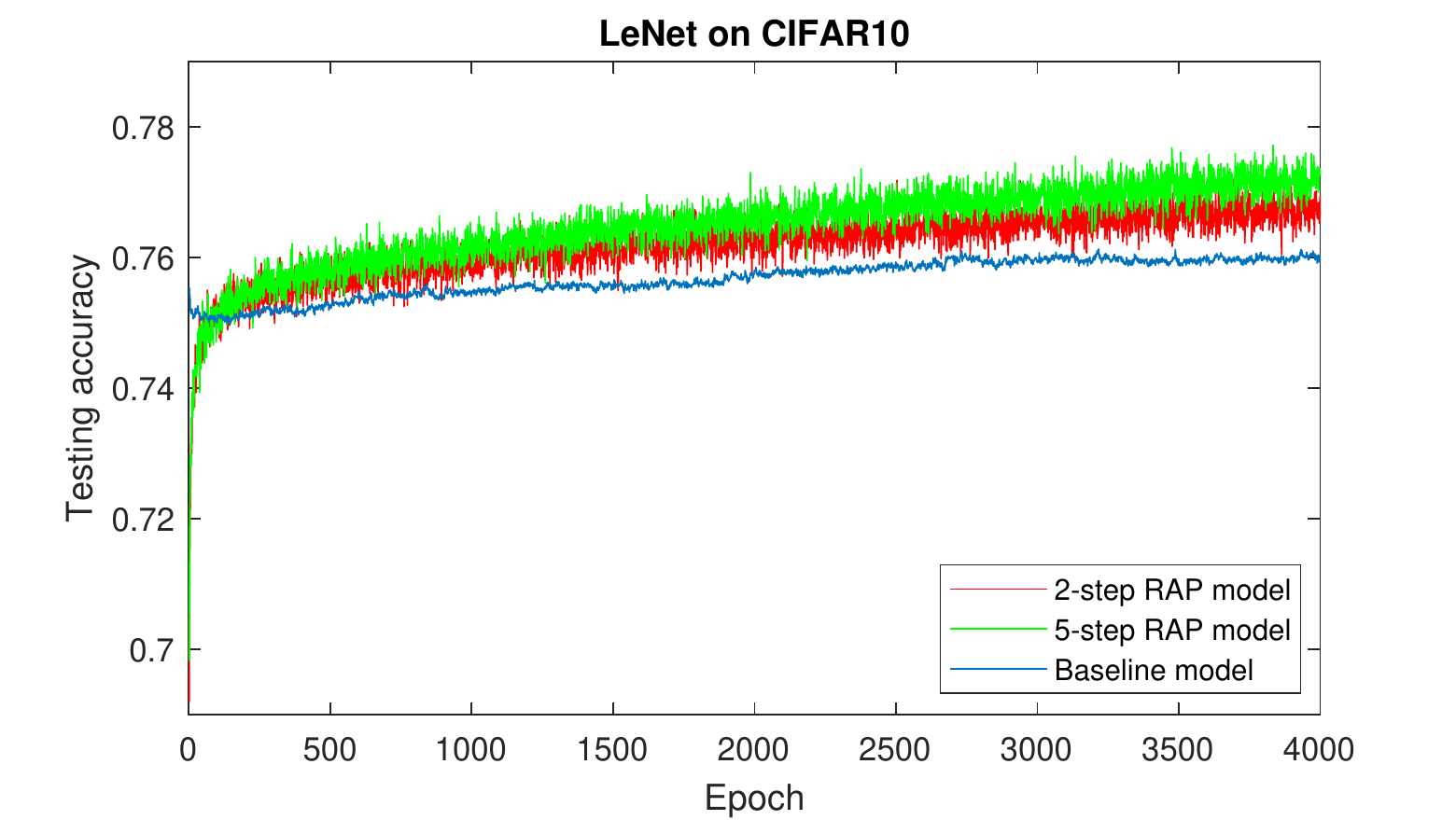}
	\includegraphics[width=7.5cm]{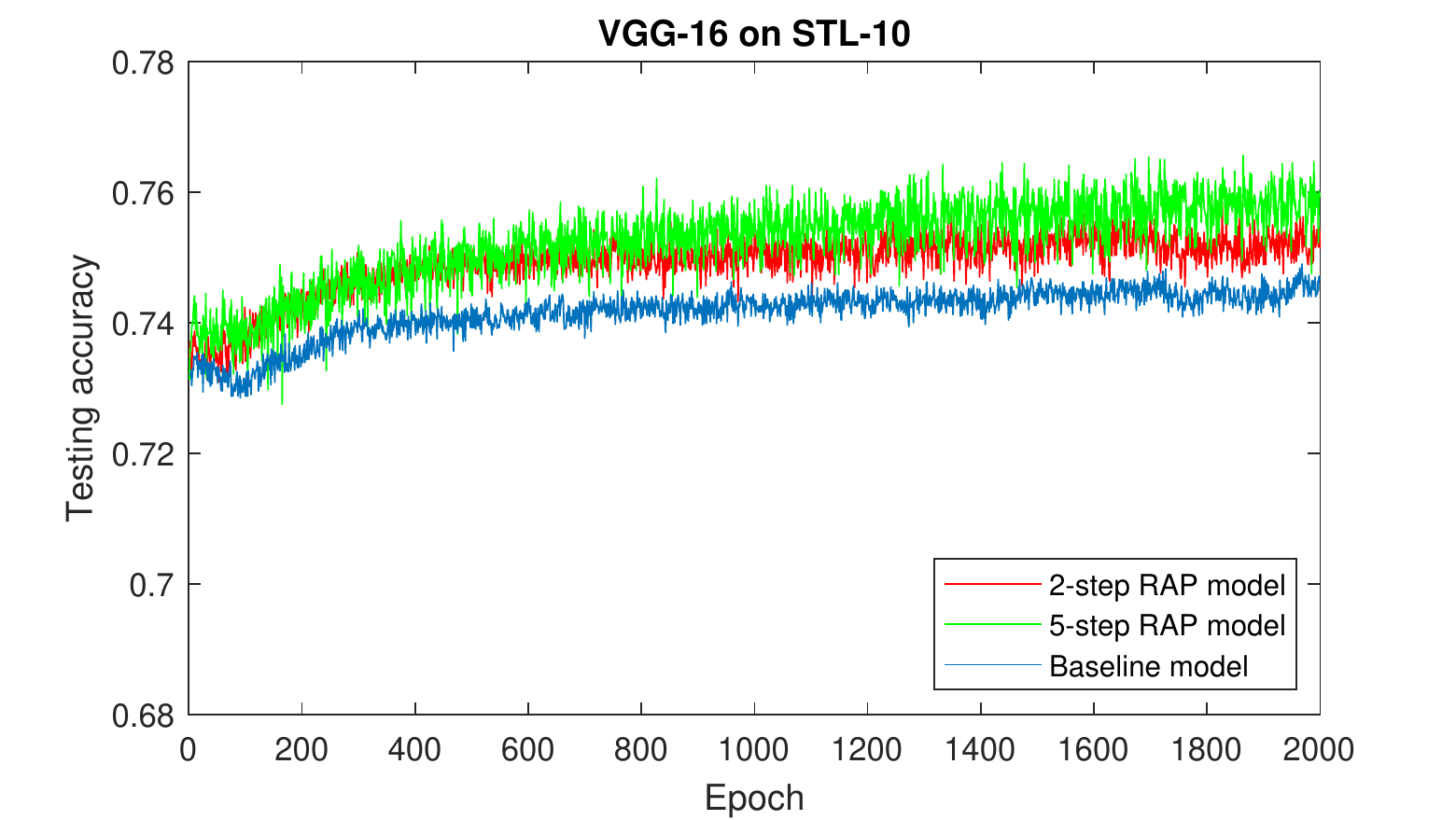}
 	\caption{The recorded accuracy of image classification of LeNet on CIFAR10 and VGG-16 on STL10.}\label{recurrent}
\end{center}
\end{figure}

\vspace*{-6pt}
\paragraph{Recurrent Process.} A natural question is whether the recurrent modeling of reinforcement learning contributes to the increased performance. To provide evidence that the behavior of the sequential process has a positive effect, we perform the experiments by varying the total time step $T$. Under the same few-shot learning scenario, a 2-step model and a 5-step model are compared. We can see from Table \ref{table_recurrent} that the performance has increased nearly 1\% to 2\% with the greater number of time steps. The results suggest that our reinforced attention continuously adapt to important information of the feature maps over time. The fact that classification accuracy gains as more time steps of actions are executed reinforces the contribution and improvement attained by the learned policy. For image classification, we take LeNet on CIFAR10 and VGG-16 on STL10 as instantiations. The curves of the test accuracy are plotted in Fig \ref{recurrent}. Interestingly, the accuracy of the 2-step RAP model increases as quickly as the 5-step model in the initial epochs, but it is gradually surpassed by the 5-step model. In our setting, the reward is closely linked to the classification accuracy. Fig. \ref{recurrent} also suggests that the learned policy maximizes the reward. The experiments strongly indicate that reinforcement learning can effectively learn from experience in this context, and that it benefits from a greater number of iterations. 

\begin{table}
\begin{center}
\resizebox{.47\textwidth}{!}{
\begin{tabular}{c|c|ccccc}
\hline

&\multicolumn{2}{c}{}  &\multicolumn{2}{c}{\textbf{5-way Acc.}}   \\  \hline

\textbf{Model} &\textbf{$T$} &\textbf{Backbone} &\textbf{1-shot} &\textbf{5-shot}  \\ \hline\hline

\rowcolor{Gray}&2 &Conv-4     &60.15$\pm$0.71   &75.76$\pm$0.50   \\

\rowcolor{Gray}&5 &Conv-4     &61.49$\pm$0.70   &77.15$\pm$0.50   \\

\rowcolor{Gray}&2 &Conv-6     &68.33$\pm$0.70   &80.43$\pm$0.45     \\ 

\rowcolor{Gray}\multirow{-4}*{\textbf{RAP-MAML}} 
&5 &Conv-6                    &69.95$\pm$0.68   &81.48$\pm$0.44      \\ \hline

\rowcolor{Gray}&2 &Conv-4      &54.01$\pm$0.65   &78.12$\pm$0.45   \\

\rowcolor{Gray}&5 &Conv-4      &56.71$\pm$0.66  &78.70$\pm$0.44   \\

\rowcolor{Gray}&2 &Conv-6      &66.27$\pm$0.68  &82.58$\pm$0.42   \\

\rowcolor{Gray}\multirow{-4}*{\textbf{RAP-ProtoNet}}  
&5 &Conv-6     &67.79$\pm$0.66  &83.78$\pm$0.41  \\ \hline

\end{tabular}} 
\end{center}
\caption{Comparisons among RAP models with different time step $T$ on CUB-200-2011.} \label{table_recurrent}
\end{table}

\section{Conclusion}
In this paper, we propose RAP, a novel attention design which is able to fit nicely into most few-shot learning baselines to refine the feature map towards a more discriminative representation. To the best of our knowledge, our approach is the first attempt to address few-shot learning via applying an attention mechanism  trained by reinforcement learning. We show that the reward design based on validation data enables the model to learn a more diverse distribution. More importantly, the key idea underlying our method is that the recurrent formulation of reinforcement learning has the nature of locating the available information from experience over time and thus enables the agent to make more accurate decisions.

\section*{Acknowledgements}
Tong Zhang is partly supported by the Swiss National Science Foundation via the Sinergia grant CRSII5-180359. We would like to thank the anonymous reviewers for their useful feedbacks.

{\small
\bibliographystyle{ieee_fullname}
\bibliography{egbib}
}

\end{document}